\documentclass[10pt,twocolumn,letterpaper]{article}

\usepackage{cvpr}
\usepackage{times}
\usepackage{epsfig}
\usepackage{graphicx}
\usepackage{amsmath}
\usepackage{amssymb}
\usepackage[belowskip=-10pt,aboveskip=0pt]{caption}

\usepackage[pagebackref=true,breaklinks=true,letterpaper=true,colorlinks,bookmarks=false]{hyperref}

\cvprfinalcopy 

\def\cvprPaperID{****} 

\ifcvprfinal\fi
\begin{document}

\title{Hybrid Style Siamese Network: Incorporating style loss in complementary apparels retrieval}

\author{Mayukh Bhattacharyya\\
{\tt\small mayukhbhattacharyya18@gmail.com}
\and Sayan Nag\\
{\tt\small nagsayan112358@gmail.com}
}

\maketitle

\begin{abstract}
   Image Retrieval grows to be an integral part of fashion e-commerce ecosystem as it keeps expanding in multitudes. Other than the retrieval of visually similar items, the retrieval of visually compatible or complementary items is also an important aspect of it. Normal Siamese Networks tend to work well on complementary items retrieval. But it fails to identify low level style features which make items compatible in human eyes. These low level style features are captured to a large extent in techniques used in neural style transfer. This paper proposes a mechanism of utilising those methods in this retrieval task and capturing the low level style features through a hybrid siamese network coupled with a hybrid loss. The experimental results indicate that the proposed method outperforms traditional siamese networks in retrieval tasks for complementary items.
\end{abstract}

\section{Introduction}

Fashion e-commerce is a booming industry right now. According to studies, the industry is \$500 billion strong. A big factor which has led to its growth and success is the large inventory it can offer to the customers. As an unwanted by-product of this advantage is the problem of plenty. How to effectively select and present the items which the customer will prefer to buy has been a problem. The solution to this, recommendation systems, has been mostly content based where informations like brands, colors and other metadata are used to recommend new unseen apparels. Latest works on this problem have been using image retrieval techniques to find out similar items from the pool of all items. For example, if an user looks at a pink striped shirt, the system will most likely offer the user other variations of pink or striped apparels. That may be a very good solution for choosing same type of apparel but does not work well if the user is looking for complementary apparels. In the above case, a pink or striped trouser may not be the best choice of style to go with a pink striped shirt.

Very few studies have approached the problem of finding the complementary clothing of a certain apparel. The problem is subjective to some extent since the sense of style and fashion varies from person to person. This paper tries to form a general sense of this style from combinations of apparels worn by people in real life and proposes and evaluates techniques based on their retrieval of complementary apparels.

The contribution of this paper is two-fold:
\\1.	It proposes a Hybrid Siamese Network model which takes into account the style information of the input images to better aid in retrieval of  complementary fashion items.
\\2.	It introduces a style loss function suited to image retrieval applications which coupled with traditional triplet loss leads to better performance in complementary items retrieval.

\section{Related Work}

Siamese Network was introduced Bromley et. al. \cite{siamese-first} and made famous by Chopra et al. \cite{Chopra}. In the recent years, it is being extensively used for image retrieval.
Although the main idea of Siamese Networks was that images with a degree of similarity in a certain context will lie closer in a latent space, it was shown in recent years that Siamese Networks can indeed be used in context of compatibility also. Bell and Bala \cite{bell-bala} first explored the stylistic similarity of items in images through Siamese Networks. The idea was extended to the domain of clothing style compatibility by Veit et al. \cite{veit}. Veit et al. first tackled the problem of retrieval of clothes based on compatibility in style instead of retrieval of visually similar items. It is the closest work to our paper in this domain of research. They based their experiments on data collected from Amazon.com and created positive and negative pairs based on user interests inferred from their activities. 
Although after \cite{veit}, quite a few work has been done in this domain, such as \cite{type-aware}; those are variations and modifications of \cite{veit} and mostly evaluate on the same dataset as \cite{veit}. Vasileva et al. in \cite{type-aware} used type-aware latent transformations to better model the Siamese networks representation of items of different types.
In Gatys et al. \cite{gatys}, neural style transfer was introduced which aimed at creating artificial artistic images by imposing a style image over a content image. It achieved its by optimizing a loss function comprised of a content loss and a style loss. The work in our paper follows the Siamese structure of \cite{veit} but incorporates secondary ideas from this work \cite{gatys} in regards to estimating the style loss part of the hybrid loss.

\section{Methodology}

\subsection{Siamese Networks}

Siamese Networks \cite{siamese-first}\cite{Chopra} involve two or three identical networks which share their weights between them. Those can be tied at the top by a single dense layer or can be used separately to obtain n-dimensional embeddings of the images in a latent space, with a distance metric between the embeddings being used as the loss.

\subsection{Hybrid Style Siamese Network}

Our modified Siamese Network follows the same basic backbone of any Siamese Network. Over that, we add a style network to implement our style loss. As can be seen in fig[1], outputs from the 4 initial convolutional feature layers are taken out and each such output is passed through a style network to produce 4 auxiliary outputs which are then ingested by our style loss. 
The choice of using 4 initial layers for auxiliary outputs is found to perform best for VGG16. However it may be different for other backbone architectures. The auxiliary outputs are used only in the style loss.

\subsection{Style Network}

Each chosen initial layer feature maps are passed through a style network. The style network first applies a batch normalization layer over the incoming feature map. After which, a gram-matrix of the features is computed in-network which is unrolled into a dense fully connected layer and then provided as an auxiliary output of the model. The gram-matrix and formulations are explained in section 3.5.

The batch normalizations are critical for the style features since it limits them to a defined feature space, thus ensuring that the auxiliary outputs remain within bounds. In absence of batch normalization, the style component of the loss goes out of bounds very quickly and thus could not have been used as a loss function.

A slightly different network structure is also possible where the batch normalization is applied after computing the gram-matrix layers. However the previous structure is favoured because of better performances.

\begin{figure*}[h]
\begin{center}
\makebox[\textwidth]{\includegraphics[width=0.7\linewidth]{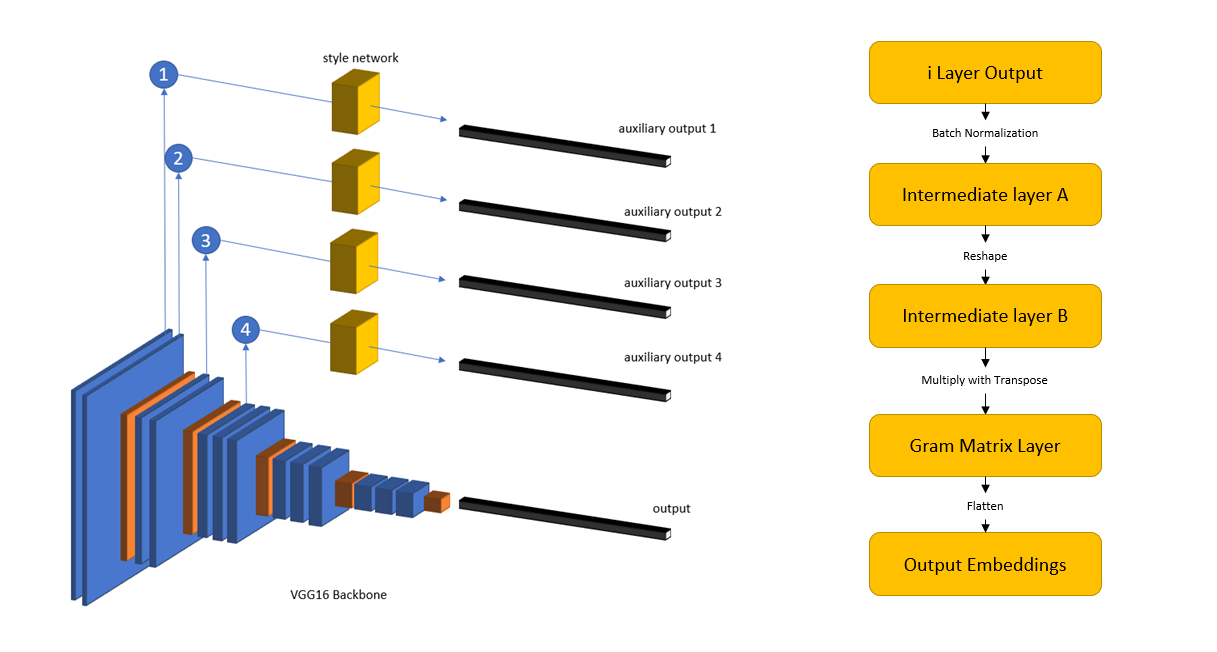}}
\end{center}
   \caption{(a) The complete Hybrid Style Siamese Network using the VGG16 backbone. Here the initial layers chosen are the 1st, 2nd, 3rd and 5th ones. (b) The Style Network architecture.}
\label{fig:short}
\end{figure*}

\subsection{Triplet Loss}

Triplet loss was introduced in \cite{triplet} and involves calculating the loss among 3 images, namely anchor image A, positive image P and negative image N. (A,P) form a positive pair (conveying a sense of similarity in most use cases) while (A,N) form a negative pair (conveying a sense of dissimilarity). In general, triplet loss combines 2 distance metrics; one between A and P, d(A,P) and the other between A and N, d(A,N).

Triplet loss strives to decrease the distance \(d(A,P)\) and increase \(d(A,N)\). A certain margin \(\alpha\) is set to separate the two distances with the positive pair distance being lower.

\begin{equation}
    L(A,P,N) = max( \alpha + d(A,P) - d(A,N), 0)
\end{equation}

In our case here, (A,P) will convey a compatible pair whereas (A,N) will convey a non-compatible pair. 

\subsection{Style Loss}

Style information from a layer is measured as the amount of correlation present between features maps of the channels in the given layer. This correlation information is computed through a gram-matrix calculated from the feature maps.

\begin{equation}
    GM(x)_{ij}^{l} = \sum_k F(x)_{ik}^{l}F(x)_{kj}^{l}
\end{equation}

Gatys et. al. \cite{gatys} defines the gram-matrix $GM$ for layer $l$ of an image $x$ such that $GM(x)_{ij}^{l}$ is the inner product between the vectorized feature maps of channel $i$ and $j$ respectively.

\begin{equation}
    L_s^{l} = \frac{1}{4n_l^{2}m_l^{2}} \sum_{i,j} ( GM(G)_{ij}^{l}  -  GM(S)_{ij}^{l})^2
\end{equation}

The style loss is defined as the difference in the gram matrices of layers evaluated between two images. Here the style loss for layer $l$, $L_s^l$ is calculated between the images $G$ and $S$. $n_l$ gives the number of channels in the layer $l$ and $m_l$ is the $height*width$ of the channels of the layer.

\subsection{Hybrid Triplet Loss}

Combining the traditional triplet loss and style loss, a hybrid loss is utilised for the hybrid siamese network.

Contrary to the use-case in neural style transfer, we use the style loss in a different way. We use a negative style loss between two images to help the network learn low-level features that can demarcate between different visual styles of two images even though they may have the same type of content. In essence, the loss forces the distance between the gram-matrices of the two images to be as large as possible. Since while evaluating compatibility of apparels, people use style as the main yardstick, learning to identify different styles in clothings will help the model to better learn the sense of compatibility.

We derive the style loss for our case as a modification of the traditional one shown in eq.(4)

\begin{equation}
    L_s^{l}(P,N) = K - K_l\frac{1}{4n_l^{2}m_l^{2}} \sum_{i,j} ( GM(P)_{ij}^{l} - GM(N)_{ij}^{l})^2
\end{equation}

Here, the style loss is computed between the positive and negative images since they are expected to be stylistically different. $K$ and $K_l$ are constants. Based on experiments, the value of $K$ is chosen to be 2. The value of $K_l$ is chosen to be $m_l$ for a certain layer $l$. We compute this loss over the model’s initial layers’ output features gathered from the positive and negative images of the triplets. Combining the style loss with the traditional triplet loss, we thus obtain a hybrid triplet loss.

\begin{equation}
    L_h(A,P,N) = w_1 * L(A,P,N) + w_2 * \sum_l L_s^l(P,N)
\end{equation}

Hybrid triplet loss, $L_h(A,P,N)$ is thus a weighted sum of the traditional triplet loss, $L(A,P,N)$ and the sum of the style losses, $L_s^l(P,N)$ over all the chosen initial layers. Here $w_1$ and $w_2$ are the corresponding weights.


\section{Experiments}

\subsection{Dataset}

Our model is evaluated on the iMaterialist \cite{imat} dataset. It contains real life images of people wearing clothes and corresponding category and segmentation details of the clothes in the images. For the experiments, among the category combinations possible, we choose the t-shirt and skirt combination. The combination of skirts \& t-shirts was chosen because it enables the model to capture a bigger extent of the variability in style. This would have been problematic in case of other categories like trousers and pants where there are only a limited few types of designs and styles prevalent.
There are a total of 3595 images having the valid (skirt \& t-shirt) combination in the iMaterialist dataset.

\subsection{Preprocessing}

Since the dataset comprises of real life images, in order to create pairs of images (each containing an apparel/type of dress) from the main image, we need to segregate the apparel portions. The pixel level segmentation annotations are used for each apparel in an image to generate masks. The masks serve the purpose of not only capturing the specific portion of the image containing the apparel but also eliminating the surrounding scene information of the images so that it does not influence the model.

\subsection{Experimental Settings and Details}

The Hybrid Style Siamese Network is compared with the normal Siamese network described in Veit et. al. \cite{veit}. To ensure a fair comparison, both the models were implemented with a VGG16 backbone. Pretrained weights trained on ImageNet were used as initialization for all the compared models. The 1st, 2nd, 3rd and 5th layers are chosen from the VGG16 backbone for style loss computation.

\begin{figure}[h]
\begin{center}
\includegraphics[width=1\linewidth]{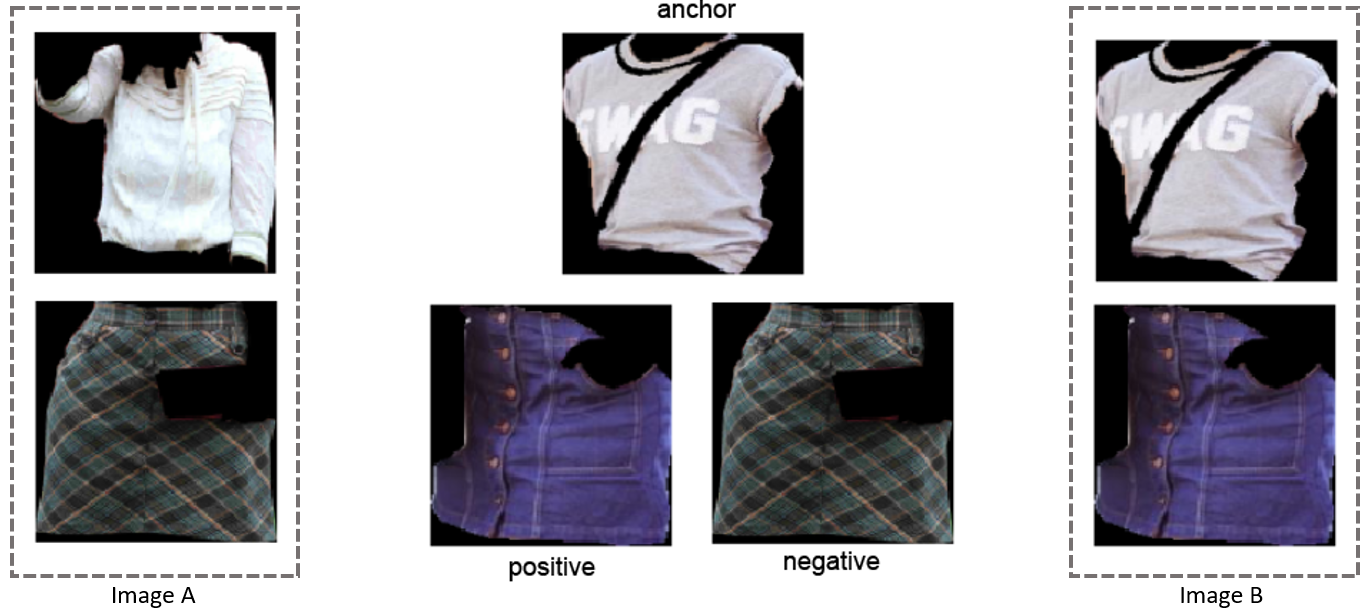}
\end{center}
\caption{Triplet formation}
\setlength{\belowcaptionskip}{100pt}
\end{figure}

Adam optimizer is used for training purposes with an initial learning rate of $8\times10^{-5}$. To create triplets, the anchor and positive samples were apparels from the same image whereas the negative one was an apparel (complementary to anchor apparel type) from a different image. Since the number of negative combinations possible is very large, the triplets are selected at random. Due to the existing randomness, in order to make the experiment robust, the experiment is carried out 3 times with 3 different random seeds and each trial is comprised of a 5-Fold cross-validation. Two learning rate decay schedules were tested and the best performing results were chosen for each of the models for each run. Experiments were run on an NVIDIA RTX 2080Ti machine with 11 GB GPU memory and 16 GB RAM.

\subsection{Evaluation Metric}

The MAP (Mean Average Precision) metric is used for evaluating the retrieval performance. During evaluation, each typeA-typeB pair (skirts and t-shirts in this case) are fed to the network and the euclidean distance is calculated. These distances are used in ranking each typeA against each typeB.

\begin{equation}
mAP = \frac{1}{2} \sum_{i}^{N} (1/R_{i}^{B} + 1/R_{i}^{A}) 
\end{equation}

$R_{i}^B$ is the rank of the typeB item of $i$th pair among all the typeB items, in terms of ascending order of distances with typeA item of the $i$th pair. $R_{i}^A$ is the opposite.

Since 5-fold cross validation was used, the MAP scores are computed on 20\% of the dataset, i.e. on 719 samples.

\subsection{Results}

The results consolidating the MAP scores from the 3 trials with the 3 different random seeds are given below.
\newline
\\
\begin{tabular}{ |p{2cm}||p{0.9cm}|p{0.9cm}|p{0.9cm}|p{0.9cm}|  }
\hline
\hline
Model&Seed 1&Seed 2&Seed 3& Mean\\
\hline
Siamese Network Veit et. al.& 0.1226&0.1323&0.1263&0.1271\\
&&&&\\
Hybrid Style Siamese Network&0.1251&0.1343&0.1329&0.1308\\
&&&&\\
Improvement (\%)&2.04&1.51&5.22&2.91\\
\hline
\end{tabular}
\newline\\
As can be seen, the Hybrid Style Siamese Network outperforms the traditional Siamese Network in all 3 of the trials. The extent of the improvement in performance varies due to the difference in randomness in the trials.

Based on the results, it can be inferred that the style network and associated style loss provides the Siamese network extra source of feature information which isn't available in normal architectures.

\section{Conclusion and Future Work}

The paper presented a novel way of encompassing the style loss of neural style transfer applications into the application of complementary apparel retrieval. The presented method gave superior results to traditional method of siamese networks and thus showed how leveraging the style information is beneficial for the retrieval of visually complementary items. 

Although the paper explores only the domain of complementary clothes retrieval, it can easily be inferred that this method will also be beneficial in general image retrieval scenarios. Since the method emphasizes on patterns at the lower level, capturing a sense of style, it can be expected that retrieval using this technique will benefit from this added advantage. 


{\nocite{*}
\bibliographystyle{unsrt}
\bibliography{egbib}

\begin{thebibliography}{1}

\bibitem{siamese-first}
Bromley et~al.
\newblock Signature verification using a "siamese" time delay neural network.
\newblock In {\em International Conference on Neural Information Processing
  Systems}, 1993.

\bibitem{Chopra}
Chopra et~al.
\newblock Learning a similarity metric discriminatively, with application to
  face verification.
\newblock In {\em IEEE Computer Society Conference on Computer Vision and
  Pattern Recognition (CVPR'05)}, 2005.

\bibitem{bell-bala}
Sean Bell and Kavita Bala.
\newblock Learning visual similarity for product design with convolutional
  neural networks.
\newblock {\em ACM Trans. Graph.}, 34(4), July 2015.

\bibitem{veit}
Veit et~al.
\newblock Learning visual clothing style with heterogeneous dyadic
  co-occurrences.
\newblock In {\em International Conference on Computer Vision (ICCV)}, 2015.

\bibitem{type-aware}
Vasileva et~al.
\newblock Learning type-aware embeddings for fashion compatibility.
\newblock In {\em Computer Vision -- ECCV 2018}, 2018.

\bibitem{gatys}
Gatys et~al.
\newblock Image style transfer using convolutional neural networks.
\newblock In {\em The IEEE Conference on Computer Vision and Pattern
  Recognition (CVPR)}, 2016.

\bibitem{triplet}
Chechik et~al.
\newblock Large scale online learning of image similarity through ranking.
\newblock {\em J. Mach. Learn. Res.}, 2010.

\bibitem{imat}
Guo et~al.
\newblock The imaterialist fashion attribute dataset.
\newblock {\em CoRR}, abs/1906.05750, 2019.

\end{thebibliography}
}

\end{document}